\documentclass[conference]{IEEEtran}
\usepackage{times}

\usepackage[numbers]{natbib}
\usepackage{multicol}
\usepackage[bookmarks=true]{hyperref}
\usepackage{graphicx}
\usepackage{subcaption}
\usepackage{booktabs}

\newcommand{\datasetName}{AVT-FCW-TrajAttn~}

\begin{document}

\title{Modeling Drivers’ Risk Perception via Attention to Improve Driving Assistance}

\author{
\authorblockN{Abhijat Biswas\authorrefmark{4},
John Gideon\authorrefmark{2},
Kimimasa Tamura\authorrefmark{2}, and
Guy Rosman\authorrefmark{2}}
\authorblockA{\authorrefmark{4} The Robotics Institute, 
Carnegie Mellon University
\authorrefmark{2} Toyota Research Institute\\
Correspondence: abhijatbiswas@gmail.com}
}
\maketitle

\begin{figure*}[ht]
    \begin{subfigure}[t]{\textwidth}
    \centering
    \includegraphics[width=0.75\textwidth]{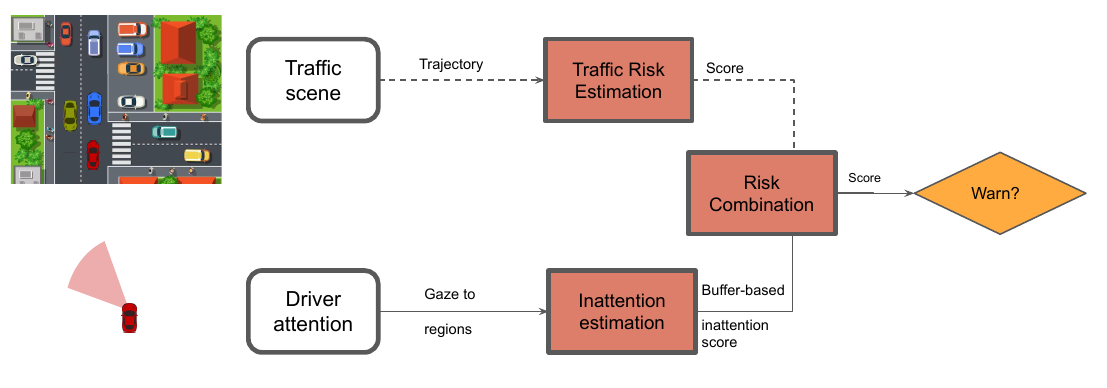}
      \caption{AttenD model (late fusion). Traffic scene risk is conceptual in~\cite{kircher2009issues}.}
      \label{fig:concept-1}
    \end{subfigure}\hfill
    \centering
    \begin{subfigure}[t]{0.75\textwidth}
    \includegraphics[width=\textwidth]{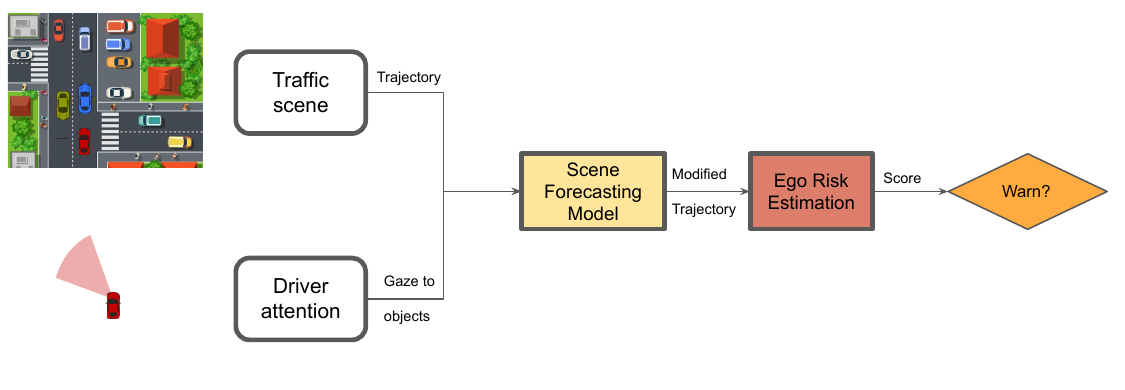}
      \caption{Our model (early fusion). The Scene Forecasting Model uses the drivers' gaze to objects to explicitly model their perceived states and hence forecasted states of other vehicles.}
      \label{fig:overall}
    \end{subfigure}\hfill
    \caption{When existing methods (a) seek to use scene risk and driver inattention in conjunction to deploy driver assistance warnings, they are usually separately computed and then combined. We propose a method to incorporate driver inattention into our a scene forecasting model that uses it to hypothesize the driver's scene perception and hence infer risk (b). This forecasting model leads to a more interpretable intermediate trajectory representation and better FCW performance on our dataset.}
    \label{fig:overall}
\end{figure*}

\begin{abstract}
Advanced Driver Assistance Systems (ADAS) alert drivers during safety-critical scenarios but often provide superfluous alerts due to a lack of consideration for drivers' knowledge or scene awareness. Modeling these aspects together in a data-driven way is challenging due to the scarcity of critical scenario data with in-cabin driver state and world state recorded together.
We explore the benefits of driver modeling in the context of Forward Collision Warning (FCW) systems. Working with real-world video dataset of on-road FCW deployments, we collect observers' subjective validity rating of the deployed alerts. We also annotate participants' gaze-to-objects and extract 3D trajectories of the ego vehicle and other vehicles semi-automatically. 
We generate a risk estimate of the scene and the drivers' perception in a two step process: First, we model the movement of vehicles in a given scenario as a joint trajectory forecasting problem. 
Then, we reason about the drivers' risk perception of the scene by counterfactually modifying the input to the forecasting model to represent the drivers' actual observations of vehicles in the scene. The difference in these behaviours gives us an estimate of driver behaviour that accounts for their actual (inattentive) observations and their downstream effect on overall scene risk.
We compare both a learned scene representation as well as a more traditional ``worse-case'' deceleration model to achieve the future trajectory forecast. Our experiments show that using this risk formulation to generate FCW alerts may lead to improved false positive rate of FCWs and improved FCW timing.

\end{abstract}

\IEEEpeerreviewmaketitle

\section{Introduction}
\label{sec:intro}

Advanced Driver Assistance Systems (ADAS) have become an integral part of modern vehicles with the objective of improving on-road safety. Some such systems are designed to alert drivers during safety-critical scenarios, such as Blind Spot Monitoring, Lane Departure Warnings or Forward Collision Warnings. However, a significant challenge faced by current ADAS implementations is the high incidence of superfluous alerts, often stemming from a failure to account for the driver's knowledge and awareness of the scene.
Repeated redundant alerts don't just annoy drivers, they risk users being desensitized to alerts or turning these off completely due to ``alert fatigue''~\cite{ancker2017effects}.

In this work we consider the issue of superfluous alerts in a particular ADAS \textemdash~ Forward Collision Warning (FCW), which are intended to alert drivers to potential frontal collisions. These systems generally rely on physics based motion models of the longitudinal motion of a lead vehicle or surface and the ego vehicle, to determine a risk estimate or collision buffer distance~\cite{chen2019vehicle, doi1994development}. While these methods sometimes account for the human drivers' reaction time as part of the calculation to determine when to warn, the humans' awareness status is not factored in and neither is the reaction time modulated based on this. This leads to redundant FCW alerts: in one study, 1873 of 2033 ($92.2\%$) deployed conventional FCW alerts were marked as ``nuisance'' or ``crash unlikely'' by observers~\cite{seaman2022evaluating}.

These elements can be challenging to model, primarily because of the scarcity of data depicting risky on-road events as well as eye gaze and driving actions. Most driving data with associated gaze tends to contain nominal driving rather than safety-critical driving~\cite{palazzi2018predicting} or was collected from in-lab video watchers rather than drivers~\cite{xia2019predicting}. As such, these data do not lend themselves to faithful modeling of complex on-road behavior, especially borne out of gaze-behavior interactions.

In this work, we explore two ways to incorporate driver attention to objects in service of improving the risk estimate that determines when to deploy an FCW warning: a learning based method that incorporates counterfactual reasoning to model human drivers' lack of attention and a non-learned method that uses similar reasoning to augment conventional FCW methods.

For the former, we use a learned trajectory forecasting method to represent the future of the scene. The inputs of other vehicle states to this method are modified when driver has not attended to those vehicles using a constant velocity counterfactual. 

In the latter formulation, a similar constant velocity counterfactual formulation is used to compute the drivers' perceived risk profile and its difference from the true risk profile is used for alert generation.

For evaluation, we extract 3D vehicle trajectories, driver attention-to-objects as well as FCW validity from an on-road FCW deployment video dataset from the MIT Advanced Vehicle Technology (AVT) Consortium’s Field Operational study~\cite{fridman2019advanced, seaman2022evaluating}.

Our contributions are:
\begin{enumerate}
    \item driver attention-aware counterfactual-based formulation of drivers' models of other vehicles during critical scenarios, which is used to generate driver risk perception estimates
    \item dataset with trajectories, gaze-to-object annotations, and need-to-warn annotations comprising both nominal and visually distracted driving
    \item experiments showing the validity of the above formulation in a downstream FCW task
\end{enumerate}

\section{Related Work}
\label{sec:background}

To provide the appropriate background for our work, we review two broad areas of work and their intersection: automated collision warning and gaze-based driver monitoring systems. 

Even though forward-looking collision warnings have been studied since the 1990s~\cite{wilson1997forward}, and are a regular feature in many modern vehicles, rear-end collisions still constitute more than a quarter of motor vehicle crashes each year~\cite{NHTSA2023}. Broadly, conventional FCW systems use a range sensor to estimate the distance of the ego to a lead vehicle and its velocity, which can be used to calculate an estimated time to collision (TTC). Then, those quantities are used to calculate a warning distance \textemdash~ when a lead vehicle gets closer than some pre-specified measure, a warning is deployed~\cite{wilson1997forward}. There have been many modifications to this basic structure, accounting for driver reaction times~\cite{wilson1997forward}, system delays~\cite{doi1994development}, road friction models~\cite{chen2019vehicle}, lead vehicle intention estimation~\cite{yang2020forward} etc.

Driver monitoring systems are also increasingly common in modern vehicles. Generally, these systems use infrared sensors in the steering column or instrument cluster to monitor driver gaze directions to determine drivers' distraction levels but do not concurrently model the scene risk. Some examples include: Cadillac SuperCruise~\cite{cadillac_supercruise_2021} (monitor eyes on/off road time), Volvo Driver Alert Control~\cite{volvo_driver_alert_2021} (alerts if the driver is distracted or showing signs of fatigue).

Using driver gaze to understand drivers' underlying cognitive state in service of driving assistance has been studied since at least the 1960s~\cite{kaluger1969driver}. However, using these signals together with outward scene context for driver assistance is a relatively new area enabled by advances in sensor quality, form factors, improved data analytics, and onboard computation \textemdash~ with the first papers exploring this appearing in the late-2000s~\cite{trivedi2007looking}. Such initial works used signals such as gaze direction in discrete traffic-facing zones as a proxy for driver attention to determine if any traffic objects were causing distracted gaze. Other methods such as AttenD estimate a time buffer which is decremented during drivers' eyes off-road time or incremented back up when the driver has eyes on-road~\cite{kircher2009issues}. The rate of these changes are modulated by a global estimate of scene risk and/or traffic complexity. However, these global measures may not capture attention to specific traffic objects that may be safety critical such as the lead vehicle during a forward collision scenario.

Finally, counterfactual reasoning about the danger of a driving scene is a recurrent idea in driving intelligence literature. For instance, Zemni \textit{et al} explored how counterfactuals can be generated to explain scene changes necessary to shift a driving model's decision, by using object-based constraints to structure the latent space~\cite{zemni2023octet}.
Closest to our work, counterfactual forecasting has been employed to estimate the relevance of vehicles to the ego vehicles plan~\cite{gupta_object_2024}. In contrast to our work, the counterfactuals are generated by perturbing inputs to a planner by modifying the LiDAR point cloud rather than an object-based input representation predictor used by us. Additionally, modifying point clouds can result in holes, which may betray the existence of an object in the counterfactual input whereas an object-level representation is cleaner.

\section{\datasetName Dataset} 
To facilitate our study, we require a dataset which includes the ability to link driver attention to particular road agents, which also contains events where we can directly measure the outcome of their inattention (e.g. hazards).
We use raw video data from an MIT-AVT consortium study which recruited 37 participants in the Greater Boston, Massachusetts area and assigned each a semi-autonomous vehicle equipped with a FCW system~\cite{fridman2019advanced, seaman2022evaluating}. Participants drove this vehicle for 4 weeks, totaling over 2200 hours of drive time. From these drives, 2033 FCW alert episodes were extracted, each spanning $5s$ before the alert was deployed and $10s$ after (henceforth, ``FCW data''). Vehicle kinematics and raw video data were recorded for each vehicle. Video data came from four synced cameras positioned around the cabin: ``views of the driver’s face, the driver’s behavior inside the cabin, the instrument cluster, and the forward roadway''.

Given the imbalance in severity rating distribution ($92\%$ were crash unlikely), we subsampled a set of 60 of 2033 episodes in the following manner. We first considered those scenarios that were originally labeled as distracted (79 instances). We then extracted trajectories using a combination of SLAM and monocular depth estimation (explained in Sec.~\ref{sec:traj_ext}). After this step, a total of 64 instances remained after others were dropped due to tracking errors. Of these 64 instances, a final 32 instances were selected to have sufficient duration and tracking quality.
In addition to these 32 distracted instances, we drew 28 non-distracted scenarios to balance the dataset. We biased this selection to include more safety-critical scenarios from the ``Crash possible (imminent/not imminent)'' subsets.

\begin{figure*}[ht]
    \centering
    \begin{subfigure}[b]{0.48\textwidth}
        \centering
        \includegraphics[width=\linewidth]{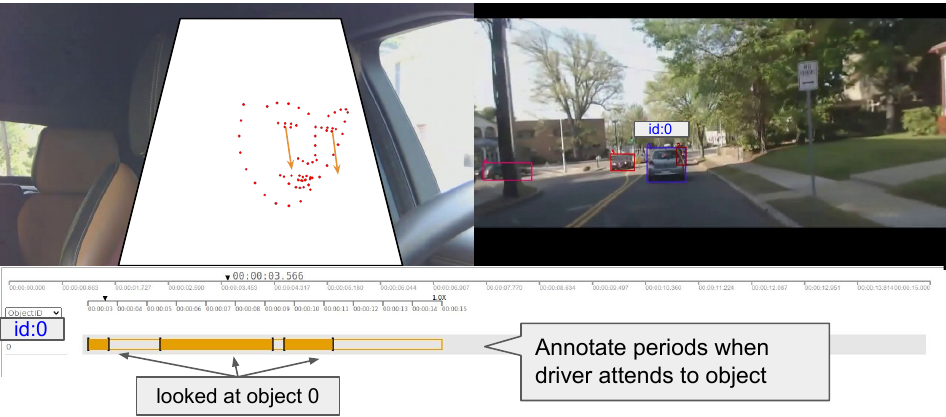}
        \caption{Observers label drivers' attention-to-objects in a temporally dense fashion of an FCW episode. Driver gaze is visible to observers (replaced with facial landmarks to protect subject confidentiality.)}
        \label{fig:attn-anno}
    \end{subfigure}
    \hfill
    \begin{subfigure}[b]{0.48\textwidth}
        \centering
        \includegraphics[width=\linewidth]{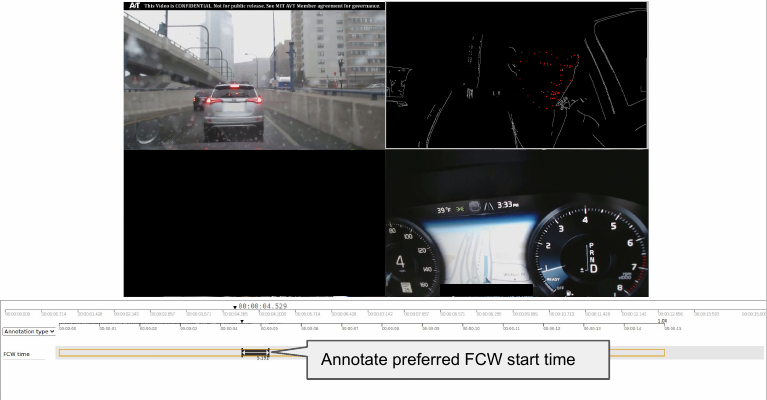}
        \caption{Binary FCW validity labels as well as FCW start times (when valid) are independently annotated by 3 observers.}
        \label{fig:fcw-anno}
    \end{subfigure}
    \caption{Annotation portals for annotating gaze-to-annotation and validity of FCW. See Sec.~\ref{sec:fcw_anno} for details}
    \label{fig:annotation}
\end{figure*}

\begin{figure}[h]
    \centering
    \includegraphics[width=\linewidth]{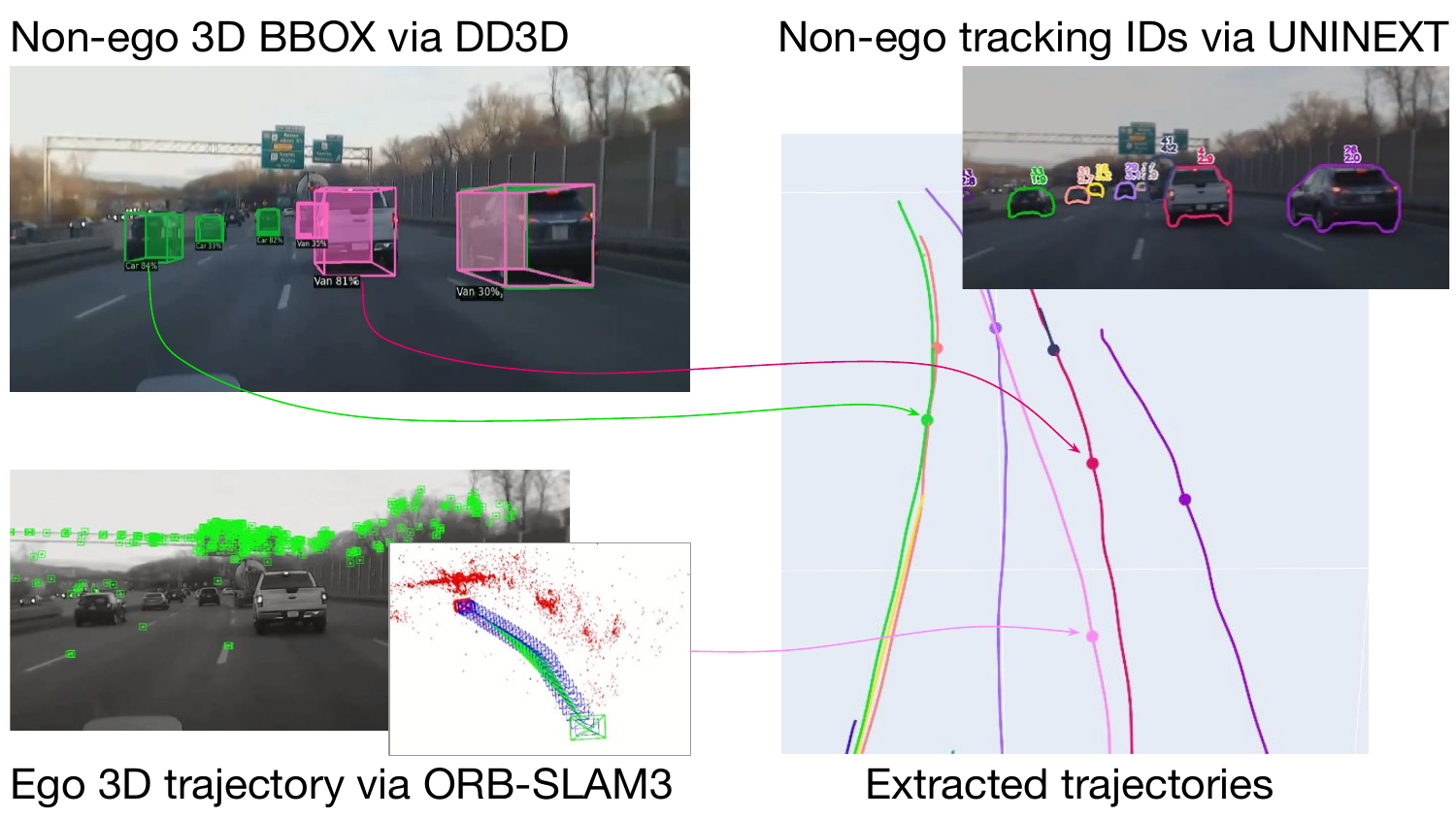}
    \caption{Trajectory extraction from road-facing video. Ego trajectory is extracted via SLAM and non-ego vehicle trajectories are extracted using off-the-shelf tracking and monocular 3D detectors. See Sec.~\ref{sec:traj_ext} for details.}
    \label{fig:traj_extraction}
\end{figure}

We used the four raw video streams to annotate driver attention, as well as the perceived necessity of the FCW deployment and the time when it should have ideally been deployed.
In particular, we used synchronized views of the drivers' face and the forward roadway with overlaid vehicle IDs to annotate drivers' attention status to important objects. This was a two step process \textemdash~ first, observers viewed $10s$ of driving video (with the FCW occurring at $5s$) and annotated the IDs of important vehicles (usually the lead vehicle in the FCW case); second, they annotated the person's gaze to each previously annotated important object during the $10s$ video. An example of the interface is shown in Fig.~\ref{fig:attn-anno}.
Finally, we also asked 3 observers to provide labels for FCW validity as well as their preferred timing of FCW if deemed necessary. While the original study had analysts assign severity ratings (``Nuisance'', ``Crash unlikely'', ``Crash possible but not imminent'', ``Crash possible and imminent''), the timing appropriateness of the FCW was not analyzed.

\subsection{Trajectory extraction description}
\label{sec:traj_ext}
The modalities required in our method include the absolute trajectories of the ego and lead vehicles in a Birds Eye View perspective. While the AVT dataset provides monocular scene videos and CAN speed profiles, it lacks LiDAR data, a common limitation in naturalistic datasets. To bridge this gap, we employed several state-of-the-art techniques. Specifically, we estimated the ego vehicle's trajectory using Simultaneous Localization and Mapping (SLAM) and combined this with positions obtained from a monocular 3D object detector to estimate the trajectories of other vehicles (example in Fig.~\ref{fig:traj_extraction}). The detailed process is outlined below:
\subsubsection{Ego Trajectory Estimation}
We used Visual SLAM to derive the ego vehicle's trajectory from scene videos. Visual SLAM constructs a map and estimates the position from videos, but it is vulnerable to dynamic objects~\cite{Saputra2018VisualSA}. Since a large number of moving vehicles are seen in the AVT dataset, we employed Masked ORB-SLAM3~\cite{masked-orb-slam3}, which ignores feature points on dynamic objects by using mask images and performs SLAM using only the static background. The mask images for dynamic objects were generated using UNINEXT~\cite{yan2023universal}, which detects and segments arbitrary objects. We masked cars, pedestrians, bikes, the dashboard at the bottom, and the AVT logo banner at the top of the images.
The intrinsic camera parameters were measured from the same camera model (Logitech C920) used in the AVT data~\cite{fridman2019advanced}. To address the scale ambiguity inherent in Visual SLAM, we scaled the odometry from SLAM to match the length of the integral of the ground truth speed profile.
\subsubsection{Other Vehicle Trajectory Estimation}
To determine the positions of other vehicles, we utilized DD3D~\cite{park2021dd3d}, a method for detecting 3D bounding boxes (cuboids) of objects from images. DD3D provided the relative 3D position, pose, and size of other vehicles relative to the ego vehicle for each frame. We projected these positions into absolute coordinates by applying the translation matrix obtained from SLAM, using the initial frame's ego vehicle position as the origin.
Since the output of DD3D also has scale ambiguity, we assumed the camera height from the ground was known for scaling. We fitted a plane to the lower four vertices of the cuboids for all detected vehicles and considered this plane as the ground. The camera height was calculated based on the specifications of the Evoque and S90 vehicles~\cite{seaman2022evaluating}. As the camera was placed close to the rear-view mirror, we assumed the distance from the top of the vehicle to the camera to be 23 cm, although this may vary depending on vehicle or setup. The vehicle trajectories were smoothed using interpolation and a Gaussian filter.

\subsection{Gaze attention annotation}

For each of the episodes in the subset, we asked observers to annotate driver attention to the most important other vehicle throughout the episode. This was done in two steps:
First, the road-facing video with bounding boxes overlaid onto each vehicle was shown to the observers and they were asked to note the ID of the object most relevant to the ego vehicle.
In the second step, we showed videos with the cabin facing image and the road facing image side-by-side (see Fig~\ref{fig:attn-anno}). Here, observers were asked to annotate the periods during which the driver was attending to the most important object marked in the previous step. 
Each episode was annotated by a single observer.

\subsection{FCW validity annotation}
\label{sec:fcw_anno}
Finally, we also ask a group of 3 observers to subjectively annotate FCW validity for each of the 60 episodes in our subset. Unlike the severity annotation in the original dataset, we explicitly ask for the observers to take into account the attention of the driver to the scene. Each observer was tasked with judging if ``FCW was required'' using in-cabin and road-facing camera videos together (see Fig~\ref{fig:fcw-anno}). To do so, they were asked to account for the ego vehicle's interactions with the other vehicle and the driver's preparedness to react.
If an observer decided that an FCW was warranted, they were also asked to annotate the time when they think the FCW should have occurred.

\section{Method}
\label{sec:method}

\subsection{Constant velocity based counterfactuals}

\begin{figure}[t]
    \centering
    \includegraphics[width=\linewidth]{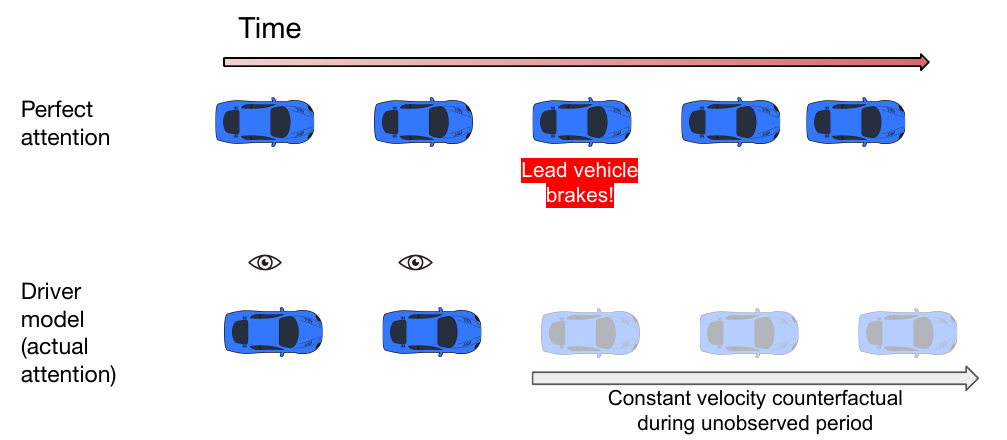}
    \caption{Our key assumption: when human drivers have not observed the lead vehicle (blue) for a while, they hypothesize its future dynamics by extrapolating a constant velocity dynamics model. This inaccurate model can lead to the drivers underestimating the scene risk, hence necessitating a warning.}
    \label{fig:counterfactual}
\end{figure}

In our method, we seek to represent that drivers often have inaccurate models of other vehicles while driving, especially stemming from relying on stale observations due to inattention or distractions. We make the assumption that when human drivers are not observing non-ego vehicles during inattentive periods (such as looking at their phone or adjusting their media console), they use past observations of those vehicles to extrapolate their motion in the future using a constant velocity assumption. Fig.~\ref{fig:counterfactual} shows a case where this kind of hypothesis can lead to dangerous driving \textemdash~ the lead vehicle slows down while the ego vehicle driver is inattentive. In this case, due to lack of information to the contrary, the ego driver assumes the lead vehicle travels at a constant velocity and hence underestimates the associated risk and continues driving according to it. 
In our methods, we attempt to account for the ego driver behaviour due to incorrect risk estimation by modifying the lead vehicle trajectories with a constant velocity forecast, at the same velocity that they were when last observed. This is a common assumption with roots in object permanence models for adults and children in psychology studies~\cite{spelke1994early}, to recent tracking literature which uses such assumptions to generate labels for occluded objects~\cite{tokmakov2021learning}.

\subsection{FCW from Learned Scene representation}

To accurately produce scene risk estimates, the learned model must model two phenomena: first, how traffic generally behaves and second, how attention (or lack thereof) impacts ego vehicle behavior. 

To model the first, we use a transformer model that forecasts the future trajectories of the ego and non-ego vehicles, given observations of their past trajectory. We could then calculate the minimum longitudinal gap in the forecast future to decide if a warning is warranted. 
However, using the true past trajectories of other vehicles may not be an accurate model of the ego driver's observations of them especially during inattentive periods. To account for this and address the second modeling desiderata, we use the counterfactual formulation shown in Fig.~\ref{fig:counterfactual}, and fill in constant velocity estimates of non-ego positions during inattentive periods. 

To generate the scene forecast, we used a version of the Planning Transformer (PlanT)~\cite{renz2022plant}. PlanT represents each scene vehicle as a token and hence lends itself to easy modifications in the input space to facilitate counterfactual reasoning based on driver attention.
Originally, the PlanT model used a representations of scene vehicles comprising their last known position and velocity rather that a subset of the past trajectory. Hence, it did not require an ego token and represented the current ego position and velocity implicitly by using it as the coordinate frame origin for all other vehicles. The original model also used a special CLS token to aggregate attention over all input tokens and produce the ego future.
Instead, we use multi-step trajectories, requiring us to use an explicit ego token and the current ego position as origin.
Finally, we predict all vehicle (ego and non-ego) trajectories jointly to encourage mutually consistent predictions.

We pre-train this transformer on the Waymo Open Motion Dataset (WOMD)~\cite{ettinger2021large}, on the joint trajectory forecasting task. Trajectory histories of $1s$ with a temporal position encoding are taken as input by the PlanT transformer to produce a scene representation per input token (vehicle). The scene representation is then decoded using a GRU to produce the future ego and non-ego trajectories. During training, a $3s$ future is used as the target while training with an L1 loss.

For the driver gaze based attention model, we use counterfactual constant velocity past inputs for the non-ego vehicles (when driver was inattentive), while the baseline learned model uses full attention for all timesteps.

\subsection{Attention-aware Conventional FCW}
In this approach, we modify a conventional FCW algorithm to account for the ego driver's attention to the lead vehicle. We use the same counterfactual assumption as before, where unobserved lead vehicle velocities are extrapolated using the last velocity observed (by the driver, see Fig.~\ref{fig:counterfactual}). 

Broadly, conventional FCW systems use physics based estimate of worst-case, immediate lead vehicle braking to determine a warning distance, $D_w$ which dictates if a warning should be deployed (if distance to lead, $D_l < D_w$). From~\cite{wilson1997forward}, the basic Stop Distance Algorithm (SDA) calculates the warning distance as $D_w$:
\begin{equation}
D_w 
= 
V_{ego} \times t_{dr} 
+ \frac{V_{ego}}{2 a_{ego}^2} - \frac{V_{lead}}{2 a_{lead}^2} 
\label{eq:dw1}
\end{equation} 
where $t_{dr}$ is the driver reaction time, $V_{i}$ and $a_{i}$ are the current velocity and the maximum deceleration of vehicle $i ~\forall~ i ~\in~ \{ ego, lead\}$.

We modify this SDA calculation to dynamically change $D_w$ based on the driver's attention periods. One straightforward way to do so would be to increase $D_w$ during inattentive periods, such that warnings are triggered at larger than usual longitudinal distances. 
However, this would not account for the duration of inattention (longer inattention periods are worse) and the relative speeds of the vehicles at the time of inattention (inattention at high speeds is worse). 
Instead, we use our constant velocity counterfactual model as described in Fig.~\ref{fig:counterfactual}, to generate counterfactual ``observation'' velocities for the lead vehicle during inattentive periods. Then, a scaled difference between the actual lead vehicle velocity and this counterfactual velocity is subtracted from $D_w$. For instance, in the case shown in Fig.~\ref{fig:counterfactual} given the counterfactual lead vehicle velocity is larger than the actual velocity, $D_w$ would increase prompting an earlier warning to the driver. Note that $D_w$ can also decrease and delay the warning if the counterfactual lead velocity is smaller than the actual velocity. This can happen when a lead vehicle speeds up during an inattention period.

\begin{equation}
D_w 
= 
V_{ego} \times t_{dr} 
+ \frac{V_{ego}}{2 a_{ego}^2} - \frac{V_{lead}}{2 a_{lead}^2} 
+ \alpha \left( {\hat{V}_{lead} - V_{lead}} \right)
\label{eq:dw2}
\end{equation} 

Here, $\hat{V}_{lead}$ represents the constant velocity counterfactual velocity. 
Empirically, we set $\alpha=1.8$.

\section{Experiments}
\begin{figure*}[t]
    \centering
    \includegraphics[width=\textwidth]{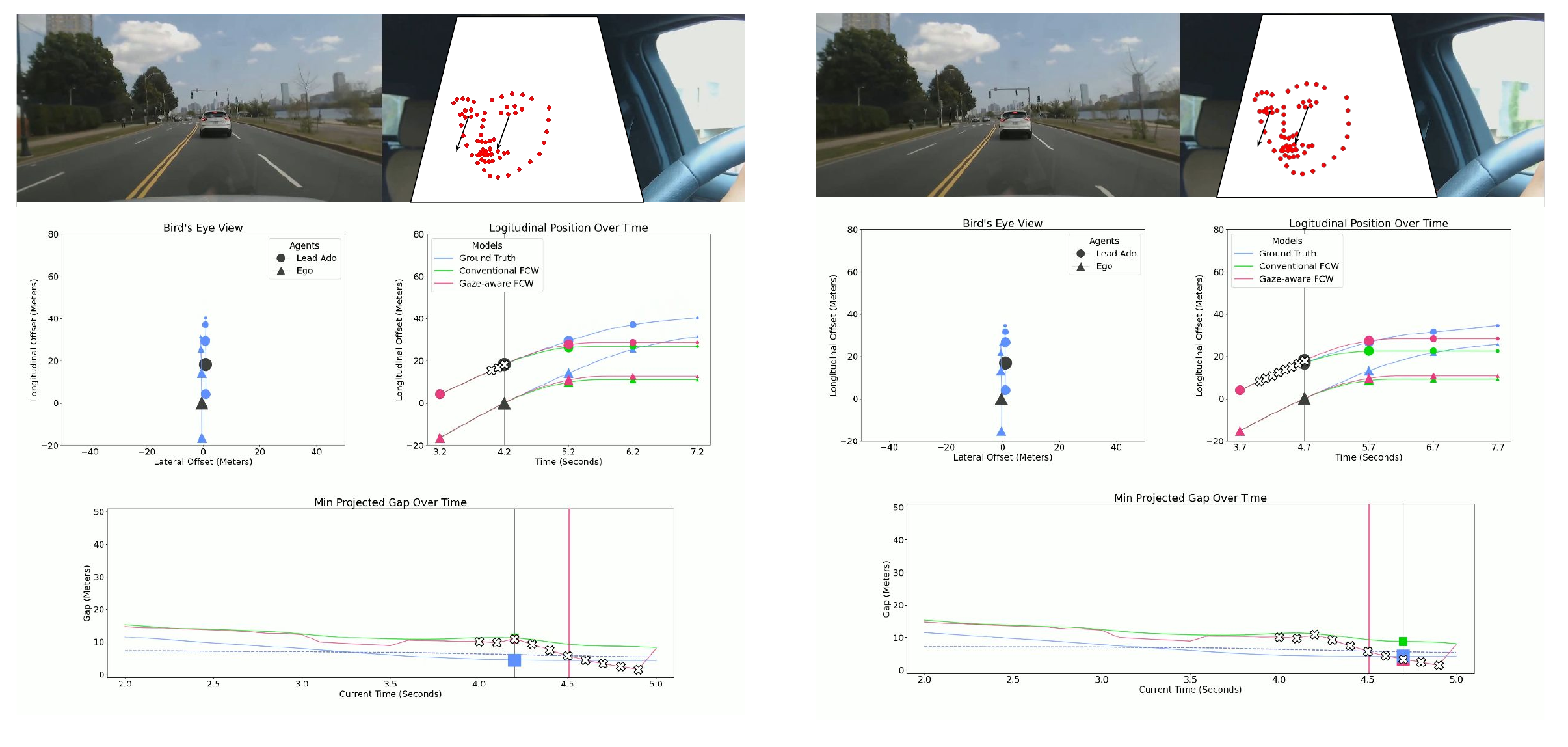}
    \caption{Results showing conventional vs. gaze aware forward collision warnings for two timesteps within the same epoch. The top left graph shows a bird's eye view (BEV) of the scene. The black icons in BEV represent the current positions of each vehicle, with the $1s$ past (behind) and $3s$ future (ahead) positions. The top right graph shows the same period, but only the longitudinal positions of each vehicle. Inattentive timesteps are represented by white cross marks. The longitudinal graph shows the actual trajectories as well as trajectories hypothesized by the Conventional and Attention-aware FCW. The bottom graph shows the minimum hypothesized future (next $3s$) gap at each timestep. With consecutive timesteps of inattention, the difference between the conventional and attention-aware lead trajectory hypotheses grows, triggering a warning at 4.5 seconds.}
    \label{fig:qual_res}
\end{figure*}

\subsection{Baseline driver assistance algorithms}
Our driver assistance baseline algorithms can be broadly classified on the basis of their awareness of the traffic scene and the human driver's attention. Typically, conventional FCW systems deployed in commercial vehicles only model the relative velocities of the ego and lead vehicles and hence fall into the traffic-aware, but attention-unaware category~\cite{mazda1994development}. On the other hand, an algorithm such as AttenD~\cite{kircher2009issues} tracks the amount of time drivers look away from an area considered ``Field Relevant for Driving'' (irrespective of the traffic scene state), issuing a driver distraction warning. This is an example of a attention-aware but traffic-unaware model. 
The AttenD model also has the capability to incorporate warning strategies based on a global measure of how critical the traffic scenario is. In this work, we use the scene risk implied by Eq.~\ref{eq:dw1} in conjunction with AttenD's driver awareness model to warn when both suggest the need for a warning.
We model the traffic scene via trajectory forecasting (using a learned or physics-based method) while also using the driver's gaze information to counterfactually model the driver's perception of the traffic scene. 
Under this taxonomy, our system and the AttenD model with critical scenario warning strategy would be classified as both traffic and gaze aware. 
We use both the gaze-only AttenD model as well as the traffic-aware version as baselines.

\subsection{Results \& Discussion}

For evaluation, we use the \datasetName dataset comprising 60 episodes of deployed FCW, evenly split between ``Crash unlikely'' and ``Crash likely'' scenarios from the original AVT consortium ratings~\cite{seaman2022evaluating}.  The FCW validity annotations described in Sec.~\ref{sec:fcw_anno} are used to evaluate the correctness of warnings emitted by each evaluated method. Since we have multiple annotators, we use a majority voting scheme where the FCW is only valid if a majority of annotators deemed it to be valid.
We compare the True Positive and True Negative rates as well as the Unweighted Average Recall across the baselines and our methods. Additionally, we seek to understand the warning characteristics of each method by comparing the FCW time buffer. This buffer is the average amount of time prior to the actually deployed FCW (from the real world data) each method suggested a warning for the driver. An example is shown in Fig.~\ref{fig:qual_res}.

We use four metrics to evaluate the performance of models, covering different aspects. 
The True Positive Rate (TPR) and True Negative Rate (TNR) comparison helps us understand how well the model  identifies when a warning is needed (TPR) and how correctly it suppresses warnings, respectively. These can also be seen as the recall value for each class (warning needed, and warning not needed).
The Unweighted Average Recall (UAR) is the average Recall acorss both classes (in this case, the average of TPR and TNR), giving us a single number balancing the two.
Finally, we use FCW buffer time as another metric of comparison, which is the earliest time at which the model would have triggered a warning. The buffer time is measured backwards from the time of warning to the $5s$ mark at each epoch. This is measured only when a warning is correctly triggered by the model.

We see that the AttenD gaze-only baseline provides 0.606 UAR performance, with balanced performance across TPR and TNR.
However, using a combination of distraction and scene risk (from SDA, Eq.~\ref{eq:dw1}) results in all but one warning being suppressed and a near-chance or trivial performance of 0.533 UAR.

Further, we see that the methods based on the learned scene representation have a TPR of 0.933. However, the TNRs are much lower at 0.065 for the normal version and 0.161 for the attention-aware version, which both indicate a very high warn rate. 
The high warn rates stem from inconsistencies in the joint predictions of the ego and lead vehicle futures. A commonly occurring pattern is a lead vehicle stop being predicted as it slows down, but no corresponding slowdown prediction for the ego vehicle. This leads to the predicted longitudinal gap unrealistically decreasing, prompting warnings more frequently than would be desirable. Given these high warn rates, the alerts generated by these models err on the side of caution and may cause alert fatigue in drivers.
We believe these errors are due to the inefficiences of learning a predictor that takes pairwise-dependent predictions well from a dataset where they are underrepresented. Handling pairwise-dependent predictions of agents is a current topic of research, and predictors are sensitive to the dataset. Due to lack of a large corpus of FCW-like pairwise trajectories, we were unable to train or fine-tune predictors on this kind of data.
Additionally, most trajectory prediction algorithms today use a variety loss~\cite{thiede2019analyzing, gupta2018social}. Under this loss, models concurrently predict several (`N') modes of behaviour and benchmarks allow a best of N evaluation~\cite{ettinger2021large}. While assumptions underlying this kind of prediction may be appropriate to encapsulate the different behaviour modes in a marginal, single-agent trajectory prediction, they are ill-suited to our case where a predictor must jointly and causally reason about the relationship between the lead and ego agents.

Next, the conventional FCW method using a stop distance calculation seems to err on the other side of the learned methods \textemdash~ producing far fewer warnings as indicated by the low 0.4 TPR and higher 0.839 TNR.
Finally, our results show that augmenting the conventional FCW with driver attention performs the best, achieving a balanced 0.733 TPR and 0.710 TNR.

The buffer time is highest for the the AttenD model, at $2.833s$ in the gaze-only case and $3s$ when accounting for scene 
risk. Since buffer time is only calculated when FCW is correctly triggered by the model, these only account for 9 and 1 correctly identified FCW-needed examples respectively.
Finally, the buffer time for the conventional and attention-aware FCW models are similar, at $1.267$ and $1.245$ seconds respectively. Our method also accounts for a higher number of FCW needed examples, 11.

The failure cases of the gaze-augmented method largely stem from false positives where the SDA is too conservative. Our dataset contains instances where observers agree a warning is warranted due to a lack of action from the driver but there is no driver inattention. In such cases, the gaze augmentation in Eq.~\ref{eq:dw2} plays no role in $D_w$ and hence, in warning calculation.

\begin{table}[tbp]
    \centering
    \begin{tabular}{llllll}
        \toprule
        Model           &  UAR & TPR & TNR & FCW Buffer \\
                        &      &     &     & $(\# correct)$ \\
        \midrule
        AttenD (gaze-only)      & 0.606 & 0.600 & 0.613 & 2.833 (9) \\
        AttenD (gaze+scene)     & 0.533 & 0.067 & \textbf{1.000} & 3.00 (1) \\        
        \midrule
        Learned (Full Attn)     & 0.499 &\textbf{0.933} & 0.065 & 0.971 (14) \\
        Learned (Driver Attn)   & 0.547 &\textbf{0.933} & 0.161 & 1.171 (14) \\
        \midrule
        Conventional FCW        &  0.619 & 0.400 & 0.839 & 1.267 (6) \\        
        Attention-aware FCW  (ours)        &  \textbf{0.722} & 0.733 & 0.710 & 1.245 (11) \\ 
        \bottomrule
    \end{tabular}
    \caption{Comparison of FCW characteristics of our method and baseline approaches. Abbreviation guide: UAR- Unweighted Average Recall; TPR- True Positive Rate; TNR- True Negative Rate.
    The Attention-aware FCW model provides the best overall true FCW detection and false FCW suppression rates. Our models show good overall (UAR) performance, suppressing false negatives (see TPR) without completely collapsing the TNR (as in the Learned models).}
    \label{tab:results}

    \vspace{-0.1in}
\end{table}

\section{Conclusion} 
\label{sec:conclusion}

In this work, we assess the benefits of incorporating a model of drivers' perceived risk in a Forward Collision Warning system. Our model uses the straightforward premise that when drivers are unable to observe vehicles on the road, they use their last seen motion to extrapolate a constant velocity. 
To allow for evaluation of models that use 3D trajectories and driver attention during safety critical scenarios, we extract these modalities from a series of monocular videos of real-world FCW deployments: the \datasetName dataset.
Using our attention-based driver perceived risk model, we show a significant reduction in the false positive rate while also improving overall FCW accuracy on \datasetName dataset. 

Our results suggest that an augmentation of existing FCW systems with driver attention and perceived risk models may lead to significantly improving false positives rates and better overall accuracy. By reducing false positives, our approach can also potentially mitigate driver desensitization to redundant alerts leading to safer on-road outcomes.
\section*{Acknowledgments}
Data for this study were drawn from work supported by the Advanced Vehicle Technology (AVT) Consortium at MIT (\url{http://agelab.mit.edu/avt}). The contents of this paper reflect the views of the authors and do not necessarily reflect those of other AVT members.

\bibliographystyle{plainnat}
\bibliography{paper_template}

\begin{thebibliography}{28}
\providecommand{\natexlab}[1]{#1}
\providecommand{\url}[1]{\texttt{#1}}
\expandafter\ifx\csname urlstyle\endcsname\relax
  \providecommand{\doi}[1]{doi: #1}\else
  \providecommand{\doi}{doi: \begingroup \urlstyle{rm}\Url}\fi

\bibitem[Aditya et~al.()Aditya, Aman, Kyle, Ping-Hua, and Zhuowen]{masked-orb-slam3}
Om~Aditya, Kushwaha Aman, Liebler Kyle, Lin Ping-Hua, and Shen Zhuowen.
\newblock Masked orb-slam3: Dynamic element exclusion for autonomous driving scenarios using masked r-cnn for increased localization accuracy.
\newblock URL \url{https://gitlab.eecs.umich.edu/v_slam/orb-slam_dynamic}.

\bibitem[Ancker et~al.(2017)Ancker, Edwards, Nosal, Hauser, Mauer, Kaushal, and the HITEC~Investigators]{ancker2017effects}
Jessica~S Ancker, Alison Edwards, Sarah Nosal, Diane Hauser, Elizabeth Mauer, Rainu Kaushal, and With the HITEC~Investigators.
\newblock Effects of workload, work complexity, and repeated alerts on alert fatigue in a clinical decision support system.
\newblock \emph{BMC medical informatics and decision making}, 17:\penalty0 1--9, 2017.

\bibitem[{Cadillac}(2021)]{cadillac_supercruise_2021}
{Cadillac}.
\newblock 2021 cadillac super cruise reference guide, 2021.
\newblock URL \url{https://www.cadillac.com/content/dam/cadillac/na/us/english/index/ownership/technology/supercruise/pdfs/2021-cadillac-super-cruise-reference-guide.pdf}.
\newblock Accessed: 2024-06-23.

\bibitem[Chen et~al.(2019)Chen, Liu, Wang, Deng, and Chen]{chen2019vehicle}
Tao Chen, Ka~Liu, Zhenyu Wang, Gang Deng, and Bin Chen.
\newblock Vehicle forward collision warning algorithm based on road friction.
\newblock \emph{Transportation research part D: transport and environment}, 66:\penalty0 49--57, 2019.

\bibitem[Doi et~al.(1994{\natexlab{a}})Doi, Butsuen, Niibe, Takagi, Yamamoto, and Seni]{doi1994development}
Ayumu Doi, Tetsuro Butsuen, Tadayuki Niibe, Takeshi Takagi, Yasunori Yamamoto, and Hirofumi Seni.
\newblock Development of a rear-end collision avoidance system with automatic brake control.
\newblock \emph{Jsae Review}, 15\penalty0 (4):\penalty0 335--340, 1994{\natexlab{a}}.

\bibitem[Doi et~al.(1994{\natexlab{b}})Doi, Butsuen, Niibe, Takagi, Yamamoto, and Seni]{mazda1994development}
Ayumu Doi, Tetsuro Butsuen, Tadayuki Niibe, Takeshi Takagi, Yasunori Yamamoto, and Hirofumi Seni.
\newblock Development of a rear-end collision avoidance system with automatic brake control.
\newblock \emph{Jsae Review}, 15\penalty0 (4):\penalty0 335--340, 1994{\natexlab{b}}.

\bibitem[Ettinger et~al.(2021)Ettinger, Cheng, Caine, Liu, Zhao, Pradhan, Chai, Sapp, Qi, Zhou, et~al.]{ettinger2021large}
Scott Ettinger, Shuyang Cheng, Benjamin Caine, Chenxi Liu, Hang Zhao, Sabeek Pradhan, Yuning Chai, Ben Sapp, Charles~R Qi, Yin Zhou, et~al.
\newblock Large scale interactive motion forecasting for autonomous driving: The waymo open motion dataset.
\newblock In \emph{Proceedings of the IEEE/CVF International Conference on Computer Vision}, pages 9710--9719, 2021.

\bibitem[Fridman et~al.(2019)Fridman, Brown, Glazer, Angell, Dodd, Jenik, Terwilliger, Patsekin, Kindelsberger, Ding, et~al.]{fridman2019advanced}
Lex Fridman, Daniel~E Brown, Michael Glazer, William Angell, Spencer Dodd, Benedikt Jenik, Jack Terwilliger, Aleksandr Patsekin, Julia Kindelsberger, Li~Ding, et~al.
\newblock Mit advanced vehicle technology study: Large-scale naturalistic driving study of driver behavior and interaction with automation.
\newblock \emph{IEEE Access}, 7:\penalty0 102021--102038, 2019.

\bibitem[Gupta et~al.(2018)Gupta, Johnson, Fei-Fei, Savarese, and Alahi]{gupta2018social}
Agrim Gupta, Justin Johnson, Li~Fei-Fei, Silvio Savarese, and Alexandre Alahi.
\newblock Social gan: Socially acceptable trajectories with generative adversarial networks.
\newblock In \emph{Proceedings of the IEEE conference on computer vision and pattern recognition}, pages 2255--2264, 2018.

\bibitem[Gupta et~al.(2024)Gupta, Biswas, Admoni, and Held]{gupta_object_2024}
Pranay Gupta, Abhijat Biswas, Henny Admoni, and David Held.
\newblock Object {Importance} {Estimation} {Using} {Counterfactual} {Reasoning} for {Intelligent} {Driving}.
\newblock \emph{IEEE Robotics and Automation Letters}, 9\penalty0 (4):\penalty0 3648--3655, April 2024.
\newblock ISSN 2377-3766.
\newblock \doi{10.1109/LRA.2024.3368301}.
\newblock URL \url{https://ieeexplore.ieee.org/document/10443043/}.
\newblock Conference Name: IEEE Robotics and Automation Letters.

\bibitem[Kaluger and Smith~Jr(1969)]{kaluger1969driver}
Nick~Alan Kaluger and GL~Smith~Jr.
\newblock \emph{Driver eye-movement patterns under conditions of prolonged driving and sleep deprivation}.
\newblock PhD thesis, Ohio State University, 1969.

\bibitem[Kircher and Ahlstr{\"o}m(2009)]{kircher2009issues}
Katja Kircher and Christer Ahlstr{\"o}m.
\newblock Issues related to the driver distraction detection algorithm attend.
\newblock In \emph{First international conference on driver distraction and inattention. Gothenburg, Sweden}, 2009.

\bibitem[{National Highway Traffic Safety Administration}(2023)]{NHTSA2023}
{National Highway Traffic Safety Administration}.
\newblock Traffic safety facts 2021: A compilation of motor vehicle crash data.
\newblock Technical Report DOT HS 813 527, National Highway Traffic Safety Administration, Washington, DC, March 2023.
\newblock URL \url{https://crashstats.nhtsa.dot.gov/Api/Public/ViewPublication/813527}.
\newblock Accessed on July 15, 2024.

\bibitem[Palazzi et~al.(2018)Palazzi, Abati, Solera, Cucchiara, et~al.]{palazzi2018predicting}
Andrea Palazzi, Davide Abati, Francesco Solera, Rita Cucchiara, et~al.
\newblock Predicting the driver's focus of attention: the dr (eye) ve project.
\newblock \emph{IEEE transactions on pattern analysis and machine intelligence}, 41\penalty0 (7):\penalty0 1720--1733, 2018.

\bibitem[Park et~al.(2021)Park, Ambrus, Guizilini, Li, and Gaidon]{park2021dd3d}
Dennis Park, Rares Ambrus, Vitor Guizilini, Jie Li, and Adrien Gaidon.
\newblock Is pseudo-lidar needed for monocular 3d object detection?
\newblock In \emph{IEEE/CVF International Conference on Computer Vision (ICCV)}, 2021.

\bibitem[Renz et~al.(2022)Renz, Chitta, Mercea, Koepke, Akata, and Geiger]{renz2022plant}
Katrin Renz, Kashyap Chitta, Otniel-Bogdan Mercea, A~Koepke, Zeynep Akata, and Andreas Geiger.
\newblock Plant: Explainable planning transformers via object-level representations.
\newblock \emph{arXiv preprint arXiv:2210.14222}, 2022.

\bibitem[Saputra et~al.(2018)Saputra, Markham, and Trigoni]{Saputra2018VisualSA}
Muhamad Risqi~U. Saputra, A.~Markham, and Niki Trigoni.
\newblock Visual slam and structure from motion in dynamic environments.
\newblock \emph{ACM Computing Surveys (CSUR)}, 51:\penalty0 1 -- 36, 2018.
\newblock URL \url{https://api.semanticscholar.org/CorpusID:3400843}.

\bibitem[Seaman et~al.(2022)Seaman, Gershon, Angell, Mehler, and Reimer]{seaman2022evaluating}
Sean Seaman, Pnina Gershon, Linda Angell, Bruce Mehler, and Bryan Reimer.
\newblock Evaluating the associations between forward collision warning severity and driving context.
\newblock \emph{Safety}, 8\penalty0 (1):\penalty0 5, 2022.

\bibitem[Spelke et~al.(1994)Spelke, Katz, Purcell, Ehrlich, and Breinlinger]{spelke1994early}
Elizabeth~S Spelke, Gary Katz, Susan~E Purcell, Sheryl~M Ehrlich, and Karen Breinlinger.
\newblock Early knowledge of object motion: Continuity and inertia.
\newblock \emph{Cognition}, 51\penalty0 (2):\penalty0 131--176, 1994.

\bibitem[Thiede and Brahma(2019)]{thiede2019analyzing}
Luca~Anthony Thiede and Pratik~Prabhanjan Brahma.
\newblock Analyzing the variety loss in the context of probabilistic trajectory prediction.
\newblock In \emph{Proceedings of the IEEE/CVF International Conference on Computer Vision}, pages 9954--9963, 2019.

\bibitem[Tokmakov et~al.(2021)Tokmakov, Li, Burgard, and Gaidon]{tokmakov2021learning}
Pavel Tokmakov, Jie Li, Wolfram Burgard, and Adrien Gaidon.
\newblock Learning to track with object permanence.
\newblock In \emph{Proceedings of the IEEE/CVF International Conference on Computer Vision}, pages 10860--10869, 2021.

\bibitem[Trivedi et~al.(2007)Trivedi, Gandhi, and McCall]{trivedi2007looking}
Mohan~Manubhai Trivedi, Tarak Gandhi, and Joel McCall.
\newblock Looking-in and looking-out of a vehicle: Computer-vision-based enhanced vehicle safety.
\newblock \emph{IEEE Transactions on Intelligent Transportation Systems}, 8\penalty0 (1):\penalty0 108--120, 2007.
\newblock \doi{10.1109/TITS.2006.889442}.

\bibitem[{Volvo Cars}(2021)]{volvo_driver_alert_2021}
{Volvo Cars}.
\newblock Driver alert control, 2021.
\newblock URL \url{https://www.volvocars.com/en-ca/support/car/xc40/article/41b16f26897fe720c0a8015138d5d35c}.
\newblock Accessed: 2024-06-23.

\bibitem[Wilson et~al.(1997)Wilson, Butler, McGehee, and Dingus]{wilson1997forward}
Terry~B Wilson, Walker Butler, Dan~V McGehee, and Tom~A Dingus.
\newblock Forward-looking collision warning system performance guidelines.
\newblock \emph{SAE transactions}, pages 701--725, 1997.

\bibitem[Xia et~al.(2019)Xia, Zhang, Kim, Nakayama, Zipser, and Whitney]{xia2019predicting}
Ye~Xia, Danqing Zhang, Jinkyu Kim, Ken Nakayama, Karl Zipser, and David Whitney.
\newblock Predicting driver attention in critical situations.
\newblock In \emph{Computer Vision--ACCV 2018: 14th Asian Conference on Computer Vision, Perth, Australia, December 2--6, 2018, Revised Selected Papers, Part V 14}, pages 658--674. Springer, 2019.

\bibitem[Yan et~al.(2023)Yan, Jiang, Wu, Wang, Luo, Yuan, and Lu]{yan2023universal}
Bin Yan, Yi~Jiang, Jiannan Wu, Dong Wang, Ping Luo, Zehuan Yuan, and Huchuan Lu.
\newblock Universal instance perception as object discovery and retrieval.
\newblock In \emph{Proceedings of the IEEE/CVF Conference on Computer Vision and Pattern Recognition}, pages 15325--15336, 2023.

\bibitem[Yang et~al.(2020)Yang, Wan, and Qu]{yang2020forward}
Wei Yang, Bo~Wan, and Xiaolei Qu.
\newblock A forward collision warning system using driving intention recognition of the front vehicle and v2v communication.
\newblock \emph{IEEE Access}, 8:\penalty0 11268--11278, 2020.

\bibitem[Zemni et~al.(2023)Zemni, Chen, Zablocki, Ben-Younes, P{\'e}rez, and Cord]{zemni2023octet}
Mehdi Zemni, Micka{\"e}l Chen, {\'E}loi Zablocki, H{\'e}di Ben-Younes, Patrick P{\'e}rez, and Matthieu Cord.
\newblock Octet: Object-aware counterfactual explanations.
\newblock In \emph{Proceedings of the IEEE/CVF conference on computer vision and pattern recognition}, pages 15062--15071, 2023.

\end{thebibliography}

\end{document}